\documentclass{article}

     \usepackage[preprint,nonatbib]{neurips_2020}

\usepackage[utf8]{inputenc} % allow utf-8 input
\usepackage[T1]{fontenc}    % use 8-bit T1 fonts
\usepackage{hyperref}       % hyperlinks
\usepackage{url}            % simple URL typesetting
\usepackage{booktabs}       % professional-quality tables
\usepackage{amsfonts}       % blackboard math symbols
\usepackage{nicefrac}       % compact symbols for 1/2, etc.
\usepackage{microtype}      % microtypography
\usepackage{makecell}
\usepackage{tabu}
\usepackage{microtype}
\usepackage{graphicx}
\usepackage{subfigure}
\usepackage{booktabs}
\usepackage{hyperref}
\usepackage{url}
\usepackage{amsthm}
\usepackage{amsfonts}
\usepackage{color}
\usepackage{multicol}
\usepackage{comment}
\usepackage{lipsum}
\usepackage{afterpage}
\usepackage{amsmath}
\usepackage{mathrsfs}
\usepackage{float}
\usepackage{listings}
\usepackage{color}
\usepackage{xcolor}
\usepackage{mathrsfs}
\usepackage{algorithm}
\usepackage[noend]{algorithmic}

\newcommand{\indep}{\perp \!\!\! \perp}

\definecolor{codegreen}{rgb}{0,0.6,0}
\definecolor{codegray}{rgb}{0.5,0.5,0.5}
\definecolor{codepurple}{rgb}{0.58,0,0.82}
\definecolor{backcolour}{rgb}{0.95,0.95,0.92}
\lstdefinestyle{mystyle}{
    backgroundcolor=\color{backcolour},   
    commentstyle=\color{codegreen},
    keywordstyle=\color{magenta},
    numberstyle=\tiny\color{codegray},
    stringstyle=\color{codepurple},
    basicstyle=\ttfamily\footnotesize,
    breakatwhitespace=false,         
    breaklines=true,                 
    captionpos=b,                    
    keepspaces=true,                 
    numbers=left,                    
    numbersep=5pt,                  
    showspaces=false,                
    showstringspaces=false,
    showtabs=false,                  
    tabsize=2
}
\newcommand*\samethanks[1][\value{footnote}]{\footnotemark[#1]}
\title{Beyond Importance Scores: Interpreting Tabular ML by Visualizing Feature Semantics}

\author{
Amirata Ghorbani\thanks{These authors contributed equally.}\\
Stanford University\\
\And
Dina Berenbaum\samethanks[1]\\
Demystify AI\\
\AND
Maor Ivgi\thanks{Corresponding authors: \url{jamesz@stanford.edu} and \url{maor.ivgi@demystify-ai.com}}\\
Demystify AI\\
\And
Yuval Dafna\\
Demystify AI\\
\And
James Zou\samethanks[2] \\
Stanford University\\
}

\begin{document}

\maketitle

\begin{abstract}
Interpretability is becoming an active research topic as machine learning (ML) models are more widely used to make critical decisions. Tabular data is one of the most commonly used modes of data in diverse applications such as healthcare and finance. Much of the existing interpretability methods used for tabular data only report \emph{feature-importance}{} scores---either  locally (per example) or globally (per model)---but they do not provide interpretation or visualization of how the features interact. We address this limitation by introducing Feature Vectors, a new global interpretability method designed for tabular datasets. In addition to providing \emph{feature-importance}{}, Feature Vectors discovers the inherent semantic relationship among features via an intuitive feature visualization technique. Our systematic experiments demonstrate the empirical utility of this new method by applying it to several real-world datasets. We further provide an easy-to-use Python package for Feature Vectors.
\end{abstract}
\section{Introduction}
\label{intro}

 As machine learning (ML) models have become widely adopted in various sensitive applications ranging from healthcare~\cite{ouyang2020video,ghorbani2020deep,esteva2017dermatologist} to finance~\cite{dixon2020machine,heaton2017deep}, the demand for interpretability has became crucial and is now even a legal requirement in some cases~\cite{goodman2017european}.
While many interpretability techniques are developed for unstructured data such as images, tabular data remains the most common modality of data in various ML applications~\cite{bughin2018notes} and the majority of computational resources are still devoted to training and deploying models trained on this form of data~\cite{jouppi2017datacenter}. Although some attempts were made to extend the impressive results of deep-learning techniques to structured data~\cite{arik2019tabnet,yang2018deep}, tree-based models consistently show superior performance~\cite{erickson2020autogluon}. Nevertheless, recent work on explainable ML has focused on improving neural network interpretability rather than tree-based models~\cite{carvalho2019machine}.

Current interpretability methods for tabular ML revolve around similar themes. These methods are either model-agnostic or specific to tree-based models and output an importance-score for each feature in the data. These scores measure either a feature's effect on predicting a single example (local) or its contribution to the model's overall performance (global). Therefore, the most informative visualization component of these methods will be a simple \emph{bar chart} of \emph{\emph{feature-importance}}{} scores. This approach, similar to the \emph{feature-importance} methods (e.g. saliency maps) used for vision and text models~\cite{ribeiro2016model,sundararajan2017axiomatic,smilkov2017smoothgrad}, is neither truly interpretable, nor actionable~\cite{poursabzi2021manipulating,kim2018interpretability}. More so, the bar chart visualization component is even less informative compared to image and text modalities.

A recent line of explainable ML research~\cite{kim2018interpretability,zhou2018interpretable,bouchacourt2019educe} on image and text data is around providing information beyond importance scores; for instance, visualizing important concepts (objects, colors, etc)~\cite{ghorbani2019towards}. Our goal is to make a similar effort for tabular ML by providing visualizations that will provide interpretations beyond simple importance scores.

\subsection{Our contribution}
We propose a new interpretability method called Feature Vectors for tabular ML. Our contribution is orthogonal and complementary to other endeavors that are focused on computing better \emph{feature-importance}{} ``scores'' i.e. scores that are more true to the model. Feature Vectors outputs an easy-to-understand visualization of the tabular dataset with two sets of information: 1) \emph{feature-importance} scores, and 2) Feature semantics. In our definition of feature semantics, two features are similar if they have similar interactions with the rest of the features in predicting the outcome and therefore, are nearly exchangeable from the model's perspective.

%The order of the section titles is: Introduction, Materials and Methods, Results, Discussion, Conclusions for these journals: aerospace,algorithms,antibodies,antioxidants,atmosphere,axioms,biomedicines,carbon,crystals,designs,diagnostics,environments,fermentation,fluids,forests,fractalfract,informatics,information,inventions,jfmk,jrfm,lubricants,neonatalscreening,neuroglia,particles,pharmaceutics,polymers,processes,technologies,viruses,vision

\begin{figure}
        \centering
        \includegraphics[width=\linewidth]{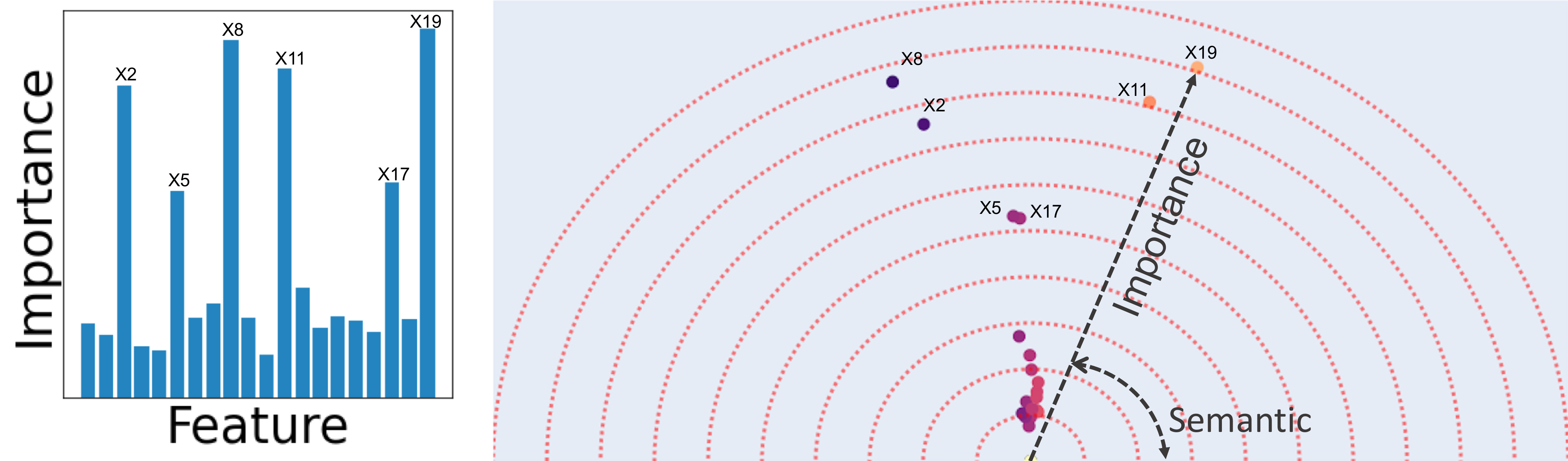} 
        \caption{\textbf{Comparison between standard feature importance scores and Feature Vector } The bar chart on the left shows the Gini importance score and the left plot shows the output of Feature Vectors. The dataset is synthetic where the features have normal distributions independent of each other. The label is generated by applying thresholds over six of the twenty features where each feature is a synonym to one other feature in that they are exchangeable in the label generation rules. The magnitude of the feature vectors shows their \emph{feature-importance}{} while the direction encodes semantics. The circles' colors also encodes angle information to make it easier to observe the similarity and dissimilarity of features.
        \label{fig:simple}}
\end{figure}

\subsection{Related Literature}

Deep learning models achieve state of the art performance in tasks related to vision, text, and speech data. Tabular data lacks spacial or temporal invariance and therefore, despite recent effort to design deep learning models specific to tabular data (e.g. TabNet~\cite{arik2019tabnet} and Deep Neural Decision Trees~\cite{yang2018deep}), tree-based models still have the strongest performance~\cite{erickson2020autogluon}. Examples of popular tree-based models are Random  Forests~\cite{breiman2001random}, XGBoost~\cite{chen2016xgboost}, Extremely Randomized Trees~\cite{geurts2006extremely}, Light GBM~\cite{ke2017lightgbm}, and CatBoost~\cite{prokhorenkova2018catboost}.

Global interpretability methods explain the model as a whole and report the importance of features for the model's overall performance. In tabular ML, a traditional approach is to compute the gain of each individual feature~\cite{loh2011classification,friedman2001elements,sandri2008bias} by computing the amount of boost in performance ("e.g. drop in training loss") gained by including a feature. Model-agnostic permutation-based methods approximate a feature's importance by randomly permuting a feature's value among test data points and tracking the decrease in the model's performance ~\cite{breiman2001random,auret2011empirical,strobl2008conditional}.

A large group of existing ML interpretability methods are developed for the local scenario where the goal is to explain the model's prediction on a single example ~\cite{lundberg2017unified, ribeiro2016should, ribeiro2018anchors, chen2018learning}. The recent focus has been on  gradient-based methods~\cite{ancona2018towards, simonyan2013deep, selvaraju2017grad} where taking the gradient of the model output with respect to the input is possible e.g. deep neural networks. Although these methods are local, by aggregating the importance of features across a large number of samples, it is possible to compute global feature importance scores. An interesting recent example is the extension of the prevalent SHAP method~\cite{lundberg2017unified} to tree-based models~\cite{lundberg2018consistent}.

 Efforts have been made to bring interpretability beyond absolute feature importance for tabular ML. RuleFit~\cite{friedman2008predictive} is a well-known algorithm which translates a tree-based model into a set of binary features and then selects the most important decision paths by training a linear classifier with $\ell_1$ regularization on top of all the binary features. A different approach is focused on computing importance scores for interaction of features rather than individual features.~\cite{tsang2020does,sundararajan2020shapley,janizek2021explaining,friedman2008predictive,hooker2004discovering,greenwell2018simple}. There have been efforts to use visualization as a better interpreatiblity tool. Partial Dependence Plots (PDP), Accumulate Local Effect plots (ALE)~\cite{apley2020visualizing}, Individual Conditional Expectation (ICE) plots and centered ICE~\cite{goldstein2015peeking}, visualize the marginal effect of one or a pair of features on the output ~\cite{friedman2001elements,zhao2021causal}. Another approach is depicting the local importance scores for a feature in a large number of individual examples in a single plot~\cite{lundberg2018consistent}. To the best of our knowledge, this is the first work that introduces a global interpretability method for tabular data that visualizes semantic explanations in addition to \emph{feature-importance} scores.

        \begin{figure}
        \centering
        \includegraphics[width=\linewidth]{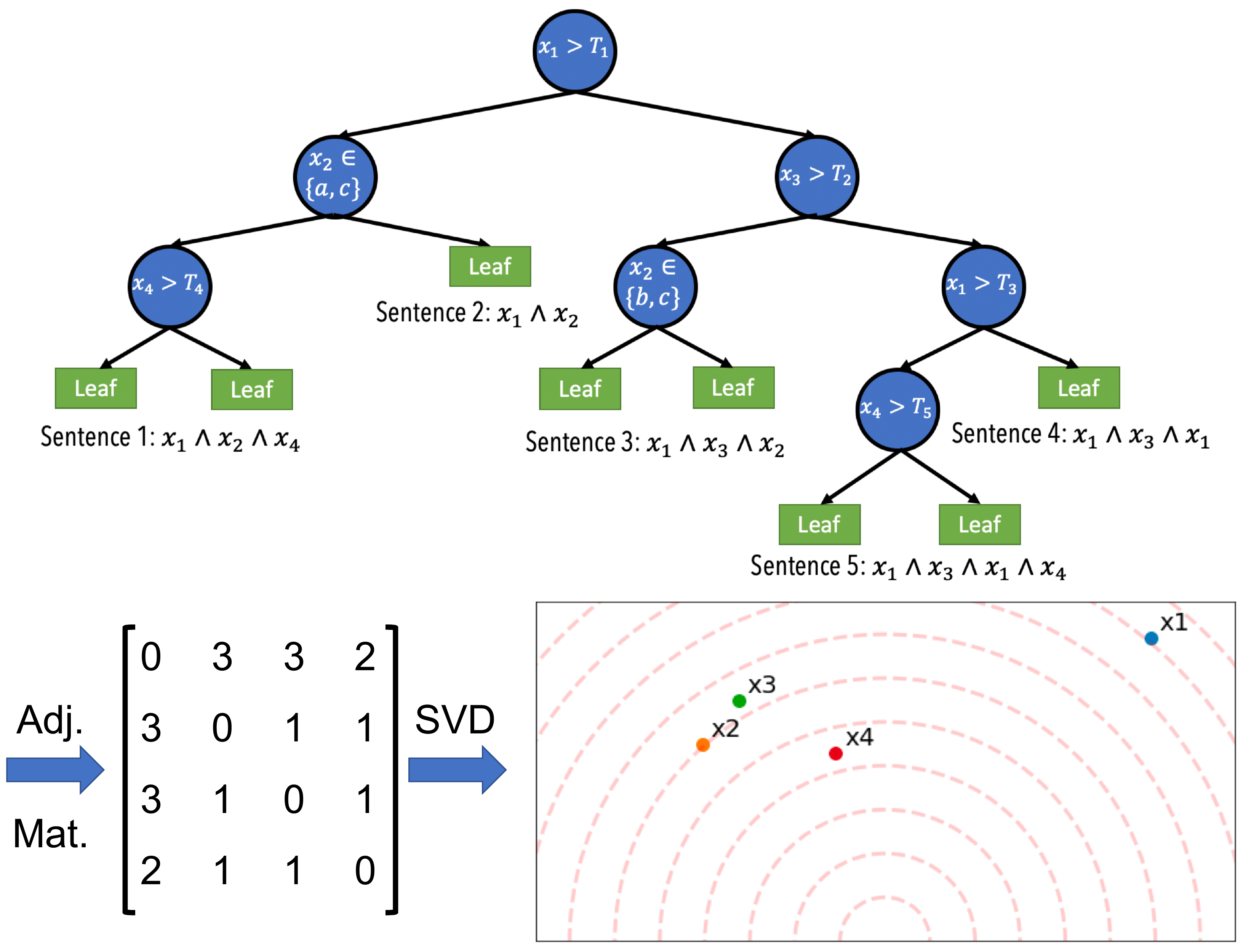} 
        \caption{\textbf{Feature Vectors} A simple description of the notion behind the Feature Vectors algorithm. The goal is to extract a meaningful embedding for each feature by looking at its co-occurrence with other features across sentences. The sentences are decision paths in a decision tree. In order to impose exchangeability among features, an ensemble of decision trees are trained such that at each split only a random subset of features is observed. After training, all sentences are extracted and the co-occurrence of each feature with other features in the sentences is computed using a sliding window. THis co-occurrence matrix is then used for extracting the embeddings.
        \label{fig:schematic}}
        \end{figure}

    \begin{figure}
    \centering
    \includegraphics[width=\linewidth]{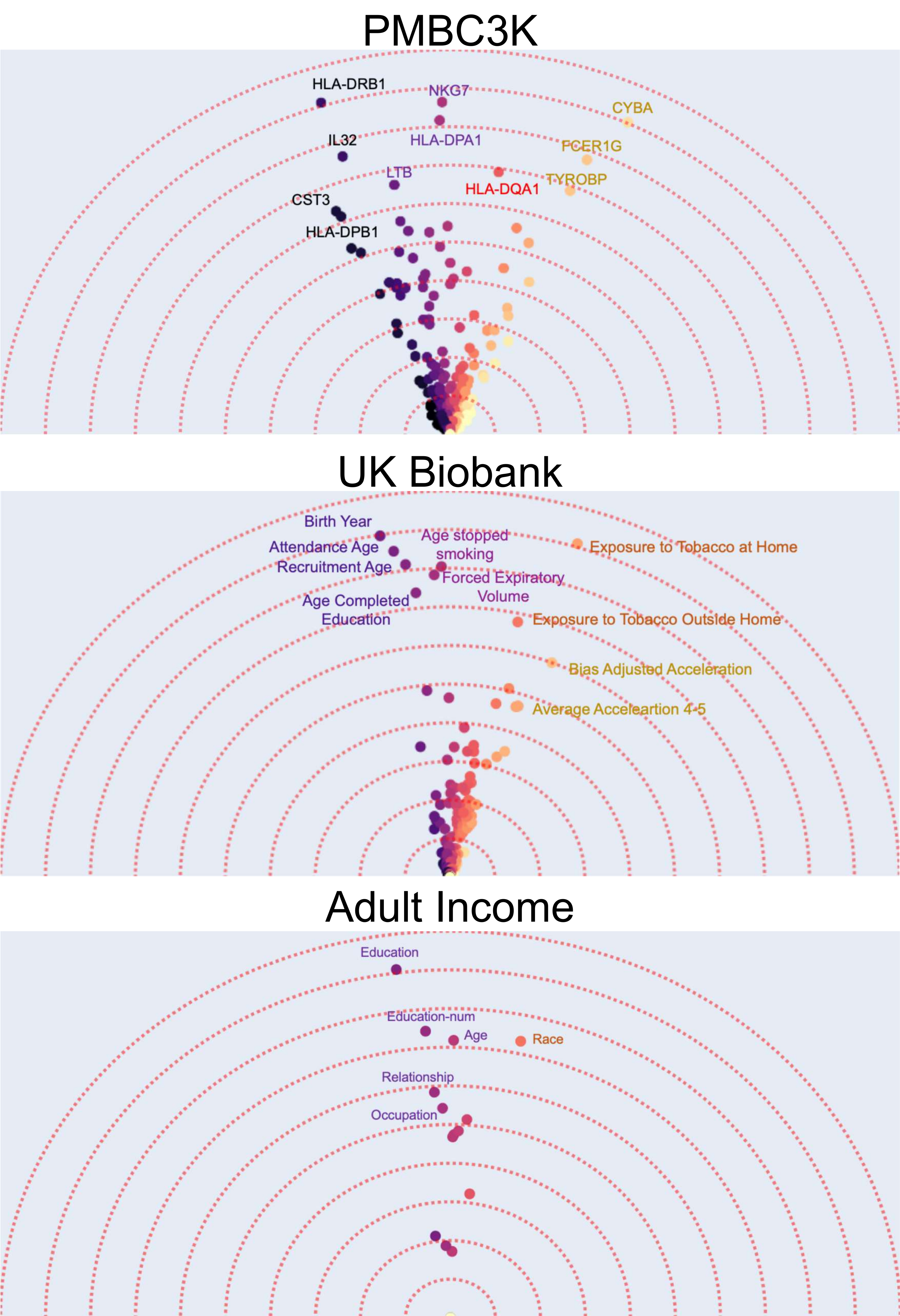} 
    \caption{\textbf{Feature Vector examples on three datasets}. Each dot corresponds to one feature. Features with similar angles are more interchangeable for the ML prediction model. For example, for predicting lung cancer in the UK Biobank, several age related features all have similar angles.  
    \label{fig:examples}}
    \end{figure}

\section{Materials and Methods}

\paragraph{\textbf{Semantic embedding}}

Our objective is to compute an embedding for each feature that contains both importance and semantic information. Inspired by word embeddings in Natural Language Processing (NLP), we are looking for a continuous dense representation of features that contains semantic information. We define two features to be semantically similar
if they are exchangeable with respect to the feature interactions that determine the model's predictions. We want this similarity to be reflected in the cosine similarity of their embeddings. Similar to the common objective in NLP, enforcing exchangeability in the context of other features will make it possible to find such a representation. In NLP, one simple embedding technique common in NLP tasks is using the co-occurrence of words in the vocabulary~\cite{bengio2003neural,collobert2008unified,mikolov2013distributed}. The intuition is that if two words are semantically similar, in a large enough corpora of sentences, their (normalized) co-occurrence frequency with other words will be similar. For a large dictionary size, in order to have a low-dimensional word-embedding, Singular Value Decomposition (SVD) dimensionality reduction is applied to the co-occurrence matrix.

There are some problems and limitations with directly applying the idea of word-vectors to features. First of all, unlike NLP where computing word embeddings is possible due to virtually unlimited supply of grammatically structured text data, in tabular data, co-occurrence of features in is not clearly defined and there is no corpus of structured co-occurrence of features. Luckily, in addition to their strong predictive performance, tree-based models naturally provide us with a structured co-occurrence of features. As shown in Fig.~\ref{fig:schematic}, a tree contains a number of decision paths where each path is a conjunction of binary conditions on a number of features. Secondly, considering a group of words with similar meanings, each word has a non-zero chance of appearing in a sentence while in a decision tree, among a group of similar features, only the most predictive feature will be chosen at its respective split. For example, consider a salary prediction task where there are two features with similar information about the outcome: \emph{earned degree} (categorical) and \emph{number of years at college} (numerical). Assuming that earned degree is a better predictor, in a given split of the tree (e.g. in a decision path that also contains \emph{age} and \emph{marital status} features), it will always be chosen over \emph{number of years at college}. As a result, it is not possible to extract the co-occurrence of \emph{number of years at college} with other features and to observe its exchangeability with \emph{earned degree}. The solution is to enforce a non-zero chance of occurrence to all features in a given decision path. The solution is to train a large ensemble of trees where each tree is only given a random subset of features (i.e. a random forest). This is the same as using a Random Forest model.

\paragraph{\textbf{Feature importance}} In natural language, word embeddings are computed with a normalization constraint so that only the semantic information is preserved in the embedding angles~\cite{levy2015improving, wilson2015controlled}. Our goal, however, is to preserve both importance and semantic information. After extracting the sentences from the trained random forest, a feature's co-occurrence with other features is a direct indicator of how many times it was chosen in a split. For example, between two similar features, the less important one will only be chosen in a split if the other feature is not present. Therefore, although the normalized co-occurrence frequency of both features is similar, the absolute co-occurrence of the less informative feature will be smaller. Thus, the Feature Vectors algorithm computes word embeddings using the counts of co-occurrence. 

\paragraph{\textbf{Human-understandable visualization}} The last desired property is to have a human-understandable visualization. After computing the co-occurrence matrix of features, to be able to visualize the features in an intuitive fashion, we reduce the dimensionality of word embeddings to two. One possible drawback is that a two dimensional embedding might lose some of the information from the feature co-occurrence pattern. In practice, for all real-world datasets that we examined, between $80\%$ to $99\%$ of the co-occurrence matrix's variance is explained by the first two components (Appendix~\ref{app:variance_ratio}).

\paragraph{\textbf{Feature Vectors Algorithm}} The pseudo-code in Alg.~\ref{alg:fv} describes the Feature Vectors algorithm's steps.

\begin{algorithm}[tb]
  \caption{\textbf{Feature Vectors}}
  \label{alg:fv}
      \begin{algorithmic}[1]
      
        \STATE {\bfseries Input:}  A dataset $X$ with $n$ data points and $d$ features (categorical or numerical) and the $n$ outputs $\mathbf{y}$ (categorical for classification and numerical for regression)
        
        \STATE {\bfseries Hyperparameters:} Number of rules $R$, co-occurrence window size $w$ 
        
        \STATE {\bfseries Output:} Feature embeddings $\mathbf{v}_1,\dots,\mathbf{v}_d \in \mathbb{R}^2$
        \vspace{1mm}
        
        \STATE {\bfseries Initializations:}
        Set of sentences $S=\{\}$
        \WHILE{$|S| < R$}
            \STATE Train a decision tree on $(X, \mathbf{y})$ where each split only observes a random subset of $\lceil \sqrt{d} \rceil$ features
            \STATE $s_t = \{\mbox{decision paths in the tree}\}$
            \STATE $S = S \cup s_t$
        \ENDWHILE
        \STATE $M \in \mathbb{R}^{d \times d}$: co-occurance of features within a window size $w$  in sentences of $S$
        \STATE  $[\mathbf{v}_1,\dots,\mathbf{v}_d]^T = $ two dimensional truncated SVD of $M$
        % \vspace{1mm}
    \end{algorithmic}
\end{algorithm}

\paragraph{\textbf{Comparison with existing methods}}
Existing global interpretability methods like gini importance, SHAP~\cite{lundberg2017unified}, and permutation importance~\cite{breiman2001random} will only provide scalar values for each feature and the result is similar to Fig.~\ref{fig:simple}'s bar chart. In comparison, the output of feature vectors, while preserving the importance information, also shows the groups of features that are semantically similar in the eyes of the model. 

Unlike Feature Vectors that computes the feature embeddings using the input-output relationship, an alternative approach is to find the feature-importance (size) and semantic similarity (angle) separately. We can use any existing importance method to find the size and then, if all the features are numerical, we can use the features' covariance matrix (instead of feature co-occurrence) to compute the semantic similarity of features. As tabular data is usually a mix of categorical and numerical variables, point-biserial correlation can be used to mitigate the issue of having mixed-type variables. We show the possible drawback of such an approach using an example in Fig.~\ref{fig:simple}. We create a synthetic dataset where the $20$ features are sampled from independent normal distributions. The label is assigned using a set of logical expressions from binary thresholds applied to the first $6$ input features e.g. $(x_1 > 2 \wedge x_2 < -3) \lor x_1 > 0$ where for each true expression, the chance of $y=1$ increases. Only $6$ features are present in the expressions and we divide them into three pairs.
The logical expressions are created such that each group of paired features are interchangeable, that is, for every expression that contains one of the features, similar expressions that contain other features of that group exist. In other words, the two features in a group are semantically similar. Fig.~\ref{fig:simple} shows that our method perfectly discovers the pairs of semantically similar features while any covariance-based method will fail as the features are independent.

\section{Experiments and Results}
    In this section, we first show a few examples of Feature Vectors implementation on widely used tabular datasets to show the usefulness of information provided by our method. The second set of experiments aims to show that the \emph{fature-importance} measure provided by this method are objectively valid. We finally use the notion of Knockoff features~\cite{candes2018panning} to examine the information provided by the angle. For all experiments, we use cross-validation for selecting the depth of trees. We empirically observe by setting $R=100000$ (number of rules), the computed feature embeddings become stable across multiple runs. We also set the window size to $w=3$.
    
\paragraph{\textbf{Easy to implement}}
    The implementation of Feature Vectors is available as a Python package and can be accessed from \url{https://github.com/feature-vectors/feature-vectors}. This tool is easy to use in general, and seamlessly integrates with Scikit-Learn's~\cite{pedregosa11a} tree-based models in a few lines of code:
\begin{lstlisting}[language=Python]
from featurevec import FeatureVec
from sklearn.ensemble import RandomForestClassifier
X, y, feature_names = data_loader()
predictor = RandomForestClassifier().fit(X, y)
fv = FeatureVec(
    mode='classify', 
    feature_names=feature_names,
    tree_generator=predictor
    )
fv.fit(X, y)
fv.plot()
\end{lstlisting}

    \begin{figure}
    \centering
    \includegraphics[width=0.7\linewidth]{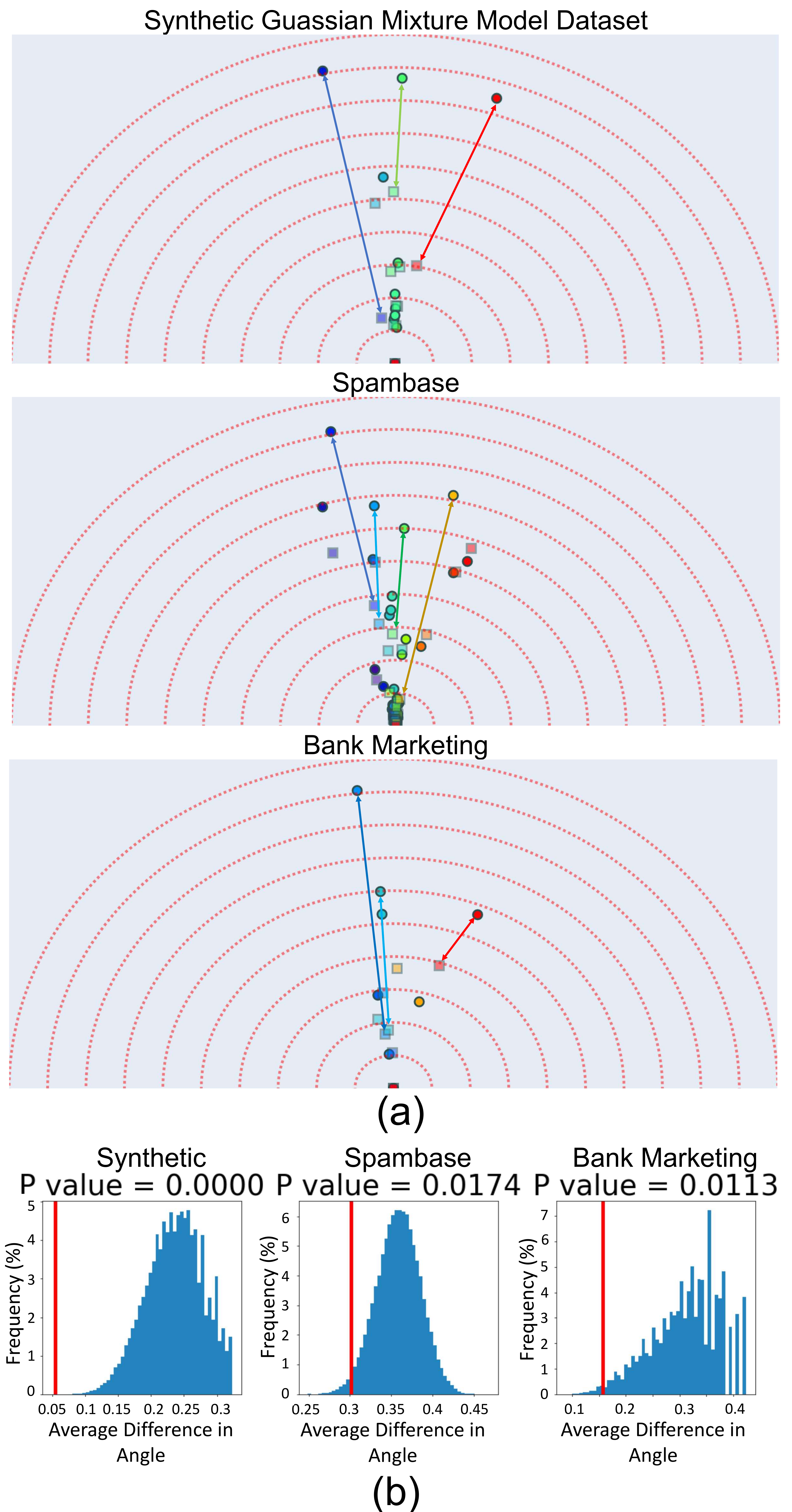} 
    \caption{\textbf{Knockoffs} (a) Feature Vector visualizes the features and their corresponding knockoffs. We can see that the feature and its knockoff have similar semantics while for non-null features, the importance (embedding size) is different. (b) A permutation test is performed. The hypothesis is that the angle difference between each feature and its knockoff is the same as angle difference between two randomly selected features. For each sample of the test, features (original and knockoffs) are paired randomly. For each pair, the difference in angle of the two features is calculated and then the difference is averaged across all pairs. By repeating this 10000 times, the blue histogram is generated. The red line shows the same average angular difference but instead of random pairing of features, each feature is paired by its knockoff. We can see that in this case, the average difference in angles is much smaller than that of random pairing and therefore, the hypothesis is rejected.
    \label{fig:kn}}
    \end{figure}

    \begin{figure}[t]
    \centering
    \includegraphics[width=\linewidth]{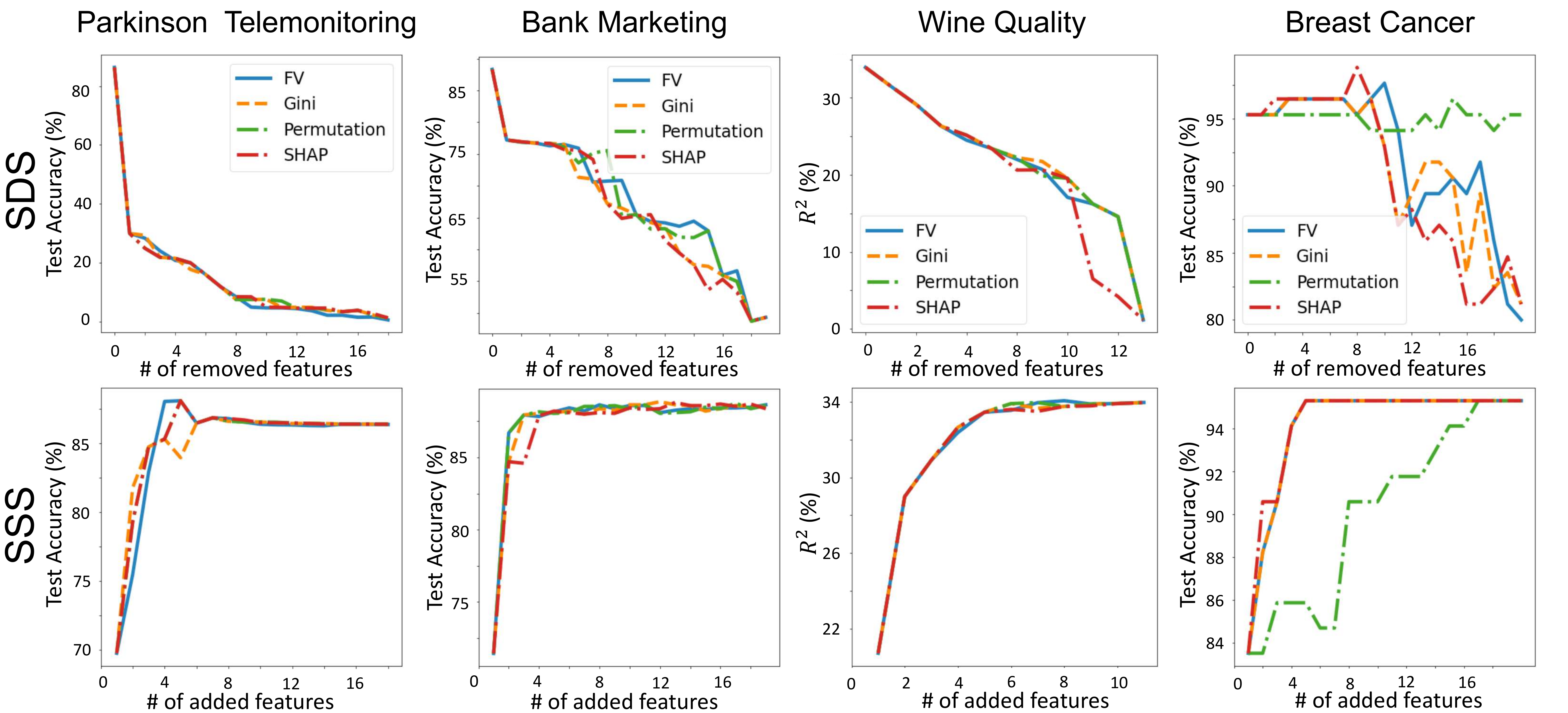} 
    \caption{\textbf{Importance scores comparison}  Feature vectors importance scores are compared with Gini, Permutation, and SHAP global importance methods. The first row shows the SDS results where we remove features from the most important and train a new model each time and report its test performance. The bottom row shows the SSS results where we add the features with the order of importance and train a new model and reports its test performance. We can see that the Feature Vectors has similar performance to other methods.
    \label{fig:drops}}
    \end{figure}

\paragraph{\textbf{Case studies on real-world datasets}}
We first apply the Feature Vectors method to three real-world datasets and discuss the explanation provided by the algorithm. Feature-vectors outputs are shown in Fig.~\ref{fig:examples} for three tabular datasets. 1) The first dataset contains the gene expression of around 1800 genes for 3000 peripheral blood mononuclear cells (PBMCs) from a healthy donor~\cite{pbmc3k}. The cells are clustered in eight different groups and the task is to determine the marker genes of the largest cluster. This is achieved by first training a binary classifier that detects if the cell is in the largest cluster by looking at the gene expression data as input features. Once the model is trained, we can look at the feature importance scores  to determine the marker genes (e.g. top-10 genes with the largest importance). Using Feature Vectors, however, in addition to finding the marker genes, we have a more detailed interpretation of how genes affect membership (Fig.~\ref{fig:examples}(a)). It can be observed that the effects of marker genes can be clustered into four major groups with similar genes. For example, although gene HLA-DQA1 does not have the highest \emph{feature-importance} score, is unique in its interaction with other genes. 2) The UK Biobank dataset has comprehensive phenotype and genotype data of $500,000$  individuals in the UK. The task is to predict if a person will be diagnosed with lung cancer in the future using $341$ phenotypic features. It can be observed that the most important features are from three main groups: A group of features that are related to age and how much time has passed since the person stopped smoking, a group of features that show the overall exposure to tobacco, and a group of features that describe the activity level. 3) The last example is the frequently used adult income prediction dataset~\cite{uci,kohavi1996scaling}. As expected, we can see that education as a categorical feature and education-num as a numerical feature are semantically similar.

\paragraph{\textbf{Validating Feature Vector angles using Knockoffs.}}
Our next experiment seeks to objectively verify the angle as a valid similarity measure. We find the notion of ``Knockoff'' features the best tool for examination. In short, for a given feature, its knockoff is a fake feature  that has perfectly similar interactions with all other features in the data. However, it only has predictive power if the original feature is removed i.e. it is conditionally independent from the outcome. Therefore, in the Feature Vectors plot, if the original feature has predictive power, we expect a feature and it's knockoff have similar angles with different magnitudes. A formal definition is as follows: for a $d$-dimensional random variable $X$, its knockoff $\tilde{X}$ is a random variable that satisfies:
\begin{itemize}
    \item Conditional independence from outcome: $\tilde{X} \indep Y | X$
    
    \item Exchangeability $\forall S \subset \{1,\dots,d\}$
    
    \[[X,\tilde{X}]_{swap(S)}\; \stackrel{d}{=}\; [X,\tilde{X}]\qquad  
\]
\end{itemize}
where $\stackrel{d}{=}$ means equality in distribution and $[X,\tilde{X}]_{swap(S)}$ means swapping the original and knockoff features in $S$. 
The exchangeability condition results in perfect semantic similarity of a feature and its knockoff while the independence condition means that if an original feature has predictive importance (i.e. it is not null: $X_j \not\!\perp\!\!\!\perp Y | X_{-j}$), its knockoff will have a smaller importance. 

Fig.~\ref{fig:kn} shows the Feature Vectors output for three datasets where the features and their knockoffs are present (the models is trained on the concatenation of [$X$, $\tilde{X}]$). The difficulty with knockoffs is that generating them for an arbitrary distribution of data is not possible. In ~\cite{gimenez2019knockoffs} it was shown that if the distribution of $X$ is a mixture of Gaussians, one can sample the knockoffs. Therefore, our first dataset is a synthetic $20$-dimensional  mixture of three Guassians, where only the first three features are non-null (predictive of the outcome). As expected, in Fig.~\ref{fig:kn}(a), we observe that the features (circles) and their knockoff (squares) have similar angles. Additionally, we can see that for non-null features, the knockoffs have a smaller importance while for other features the knockoff and the original feature have similar feature vectors. For the other two real-world datasets, the validity of the generated knockoffs depends on the validity of the Gaussian Mixture Model (GMM) assumption. Following ~\cite{gimenez2019knockoffs}, we fit a GMM model to the data. We can see in Fig.~\ref{fig:kn}(a) that the features and their knockoffs have similar angles. In addition to the subjective observation, we perform a permutation test in Fig.~\ref{fig:kn}(b) to measure the semantic similarity of original and knockoff features. The null hypothesis is that the average difference in the angle between features and their knockoffs is not different from the average difference in a random pairing of features. As expected, the hypothesis is rejected with p-values close to $0.01$.

\paragraph{\textbf{Validating \emph{feature-importance} scores}} We conduct quantitative experiments to verify the \emph{feature-importance} generated by the Feature Vectors visualization. There exist various objective metrics for measuring the goodness of global \emph{feature-importance} methods and following existing work~\cite{dabkowski2017real, ghorbani2019towards}, we will focus on two intuitive and widely used metrics:
\begin{itemize}
    \item Smallest Destroying Subset (SDS) -- Smallest subset of features that by removing them an accurate model cannot be trained.
    \item Smallest Sufficient Subset (SSS) -- Smallest subset of features that are sufficient for training an accurate model.
\end{itemize}

To measure SSS (SDS), we start removing (adding) features from the most important to the least important one by one (based on each method's importance scores) and each time train a new model and measure its performance. Fig.~\ref{fig:drops} shows the performance of Feature Vectors compared to three other famous global interpretability methods on four UCI ML repository datasets~\cite{uci}. Other methods are: 1) Gini Importance score which computes the number of times each feature has been used in a split proportional to the number of samples it splits. 2) Permutation Importance~\cite{breiman2001random} which removes features one by one by permuting its value in different data points and computes the drop in model's performance. 3) TreeSHAP~\cite{lundberg2018consistent}, which models the features as players in a cooperative game and computes the contribution of each feature to the collaboration using the notion of Shapley value. 
We can see that Feature Vectors has a comparable performance to the other three and as expected, correlates very well with the Gini importance method. More examples are provided in Appendix~\ref{app:drops}. All in all, while Feature Vectors provides feature-importance scores that are as good as other existing interpretability methods, it provides semantic information of the features.

\section{Discussion and Conclusions}

We introduce Feature Vectors, a global interpretability method for tabular ML. Feature Vectors takes a tabular dataset as an input and outputs a visualization of features where each feature is shown by its two dimensional embedding. The embeddings explain two main aspects of the data: 1) the magnitude of the embedding gives how the importance of the feature to the predictive power of the model. b) The direction of the corresponding feature shows how semantically similar or dissimilar it is to other features. Semantic similarity is defined as an interpreation tool where two features that are semantically similar have similar interactions with other features in predicting the outcome . Through extensive quantitative and qualitative  experiments, we show that using Feature Vectors can help users have a better understanding of the model while not losing any information compared to other existing methods. This work opens a new window towards creating better interpretability tools for tabular data and using the element of visualization as a tool for intuitive explanations of feature interactions. One interesting future direction is to extend this global interpretability method for local explainability as well. Feature Vectors is intrinsically limited to the tree-based models that randomly select subsets of features (e.g. random forests) and an important future direction is to extend the notion of feature semantics to other models that are used for tabular data.

\bibliography{references}
\bibliographystyle{aaai}

\newpage
\appendix
\section{Two dimensional embeddings are sufficient}
\label{app:variance_ratio}
Table~\ref{table:variance_ratio} shows the percentage of explained variance of the co-occurrence matrix using the 2-dimensional Feature Vectors embedding.

\begin{table}[t]
    \caption{\textbf{Explained Variance Ratio} \hspace{10pt}\label{table:variance_ratio} \newline}
    \centering
    {
    \begin{tabular}{ c | c}
    
            \makecell{Dataset} &
            \makecell{Explained Variance ($\%$)}\\
        \hline
        UK Biobank Lung Cancer Prediction & $81$\\
        PBMC3k & $85$\\
        Adult Income & $92$\\
        Spambase & $86$\\
        Bank Marketing & $98$\\
        Parkinson Telemonitoring & $99$\\
        Wine Quality & $98$\\
        Breast Cancer & $87$\\
        
    \end{tabular}}
\end{table}    

\section{Comparing \emph{feature-importance} methods}
\label{app:drops}

\begin{figure}[t]
    \centering
    \includegraphics[width=0.9\linewidth]{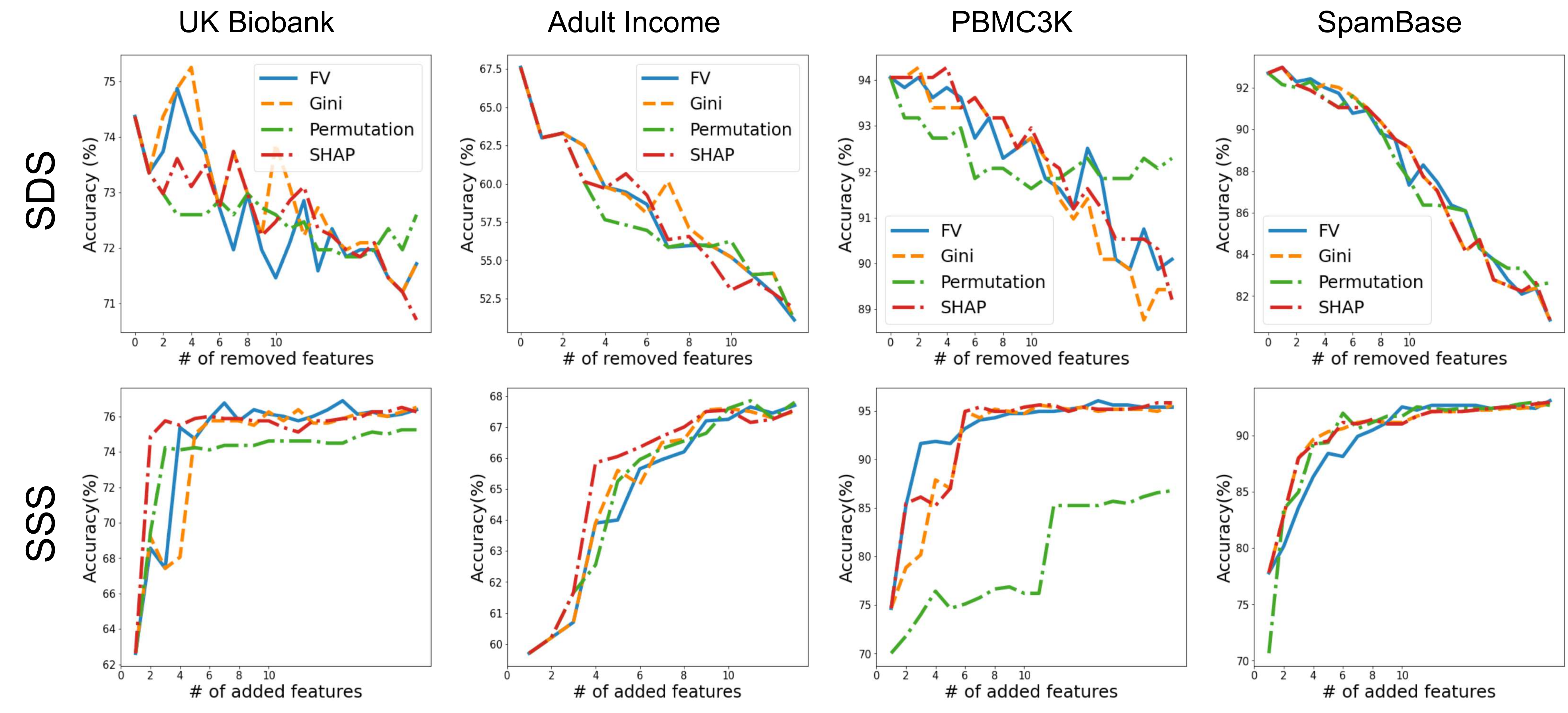} 
    \caption{\textbf{Importance scores comparison} Feature vectors importance scores are compared with Gini, Permutation, and SHAP global importance methods. The first row shows the SDS results were we remove features from the most important and train a new model each time and report its test performance. The bottom row shows the SSS results were we add the features with the order of importance and train a new model and reports its test performance. We can see that the Feature Vectors has similar performance to other methods. 
    \label{fig:app_drops}}
    \end{figure}

In Fig.~\ref{fig:app_drops}, we compare the feature importance scores of different methods using the SSS and SDS metrics for four more datasets.

\end{document}